\documentclass[letterpaper]{article} 
\usepackage[preprint]{aaai2027}  
\usepackage[hyphens]{url}  
\usepackage{graphicx} 
\usepackage{natbib}  
\usepackage{caption} 
\usepackage{algorithm}
\usepackage{algorithmic}
\usepackage{multirow}

\usepackage{newfloat}
\usepackage{listings}
\DeclareCaptionStyle{ruled}{labelfont=normalfont,labelsep=colon,strut=off} 
\floatstyle{ruled}
\newfloat{listing}{tb}{lst}{}
\floatname{listing}{Listing}

\usepackage{booktabs}
\usepackage{colortbl}   
\usepackage{enumitem}   

\newcommand{\finperma}{\textsc{FinPerMA}}
\newcommand{\impactc}{\textsc{ImpactConstraint}}
\newcommand{\eventr}{\textsc{EventReaction}}

\usepackage{amsmath,amssymb,amsthm}
\usepackage{microtype}

\title{\finperma{}: A Theory-Informed, Event-Grounded Personalized-Memory Benchmark for LLM Agents}
\author{
    Ben Wang\textsuperscript{\rm 1},
    Kang Zhou\textsuperscript{\rm 1},
    Lifan Guo\textsuperscript{\rm 1},
    Feng Chen\textsuperscript{\rm 1},
    Chi Zhang\textsuperscript{\rm 1}
}
\affiliations{
    \textsuperscript{\rm 1}Qwen DianJin Team, Alibaba Cloud Computing\\
    wangben1619@gmail.com\\
    \{wuyue.zk, lifan.lg, betterman.chenf, edward.zhang\}@alibaba-inc.com
}

\begin{document}

\maketitle

\begin{abstract}
Large language model (LLM) agents are increasingly used as personalized assistants in high-stakes domains such as financial advising, yet it remains unclear whether they can maintain and update an individualized user model over long horizons. Existing personalized-memory benchmarks primarily test factual retention or rely on weakly constrained model-generated trajectories, leaving event-driven preference adaptation underexplored. We introduce \finperma{}, an event-grounded benchmark that evaluates personalized memory against frozen longitudinal investor trajectories. Its generation pipeline combines deterministic, theory-informed impact rules, controlled LLM narration, and automated quality screening; a \emph{Post-Shock} checkpoint isolates whether an agent has integrated a material event into its persistent user model. On $2{,}994$ questions from 276 personas, seven frontier LLMs and up to seven memory configurations remain far from saturated: no full-context configuration exceeds $\approx\!0.47$ overall accuracy or $\approx\!39\%$ on multiple-choice questions. Attribution analysis shows that summary-based memory often preserves factual details while losing the preference signals needed for personalization; simple retrieval can therefore outperform purpose-built memory systems, with the gap widening after shocks.
\end{abstract}


\section{Introduction}

Recent advances in large language models \citep{brown2020gpt3,openai2023gpt4,anthropic2024claude} have enabled a new generation of agentic systems that maintain long-lived interactions with individual users \citep{park2023generative,packer2023memgpt,chhikara2025mem0}. In domains such as personal finance, medical advisory, and education, these agents are increasingly expected not merely to answer questions in isolation, but to \emph{develop} and \emph{use} a persistent, individualized model of the user across sessions spanning weeks, months, or years. A financial advisor built on top of an LLM, for instance, must recall a client's stated risk profile, revise it in light of major market events, and act on the revised profile when a new investment opportunity is proposed. The central methodological question is whether contemporary systems satisfy these expectations or merely produce shallow surface-form matches that \emph{look} like personalization.

\paragraph{Personalized memory as a benchmark target.}
A rapidly growing body of work has begun to benchmark LLMs' capacity for personalized memory. PersonaMem seeds persona attributes in multi-session dialogues and later queries them \citep{jiang2025personamem}. LoCoMo evaluates very long conversations with hundreds of dialogue turns \citep{maharana2024locomo}, while LongMemEval tests temporal reasoning, multi-session update, and knowledge synthesis \citep{wu2024longmemeval}. In parallel, several systems provide dedicated memory subsystems for LLM agents. MemGPT uses an operating-system-style hierarchy \citep{packer2023memgpt}, and MemoryBank applies an Ebbinghaus-inspired forgetting curve \citep{zhong2024memorybank}. Mem0 supports scalable extraction and retrieval \citep{chhikara2025mem0}; MemOS treats memory as a first-class OS-managed resource \citep{li2025memos}; and A-Mem uses Zettelkasten-style semantic linking \citep{xu2025amem}. Together, these benchmarks and systems constitute the empirical landscape our work engages.

\paragraph{Two gaps in current benchmarks.}
Despite substantial progress, two gaps limit our ability to test whether an agent maintains and reasons from an evolving user model.

\textbf{(G1) Limited event-conditioned state change.} Existing benchmarks emphasize factual retention, long-context recall, or explicit information updates. They provide less control over how a user's latent preferences change after consequential events, making it difficult to separate recalling the past from reasoning with an updated user state.

\textbf{(G2) Limited label auditability.} When post-event trajectories are produced by weakly constrained LLM generation, properties of the generator can become entangled with the evaluation target. \finperma{} reduces this dependence by using deterministic rules to constrain state changes, then freezing the resulting dialogues and labels before evaluating any memory system.

\paragraph{Why finance is a uniquely favourable testbed.}
Finance provides a useful testbed for controlled preference evolution. Real events with verifiable timestamps, including the 2020 COVID crash, the 2022 rate-hike cycle, and the 2023 SVB collapse, supply external timeline anchors. Behavioral-finance research identifies mechanisms and regularities that motivate interpretable operationalizations of risk response, experience, memory, and reference dependence \citep{tversky1992advances,barberis2001prospect,malmendier2011depression,malmendier2016inflation,bordalo2020memory,guiso2018time,cohn2015evidence}. Financial decisions also make stale or inconsistent user models consequential: an assistant must preserve stable constraints while adapting recommendations after material changes. Existing financial LLM resources such as FinGPT \citep{yang2023fingpt} and FinBen \citep{xie2024finben} do not center this longitudinal personalized-memory problem.

\begin{figure*}[t]
\centering
\includegraphics[width=0.86\textwidth]{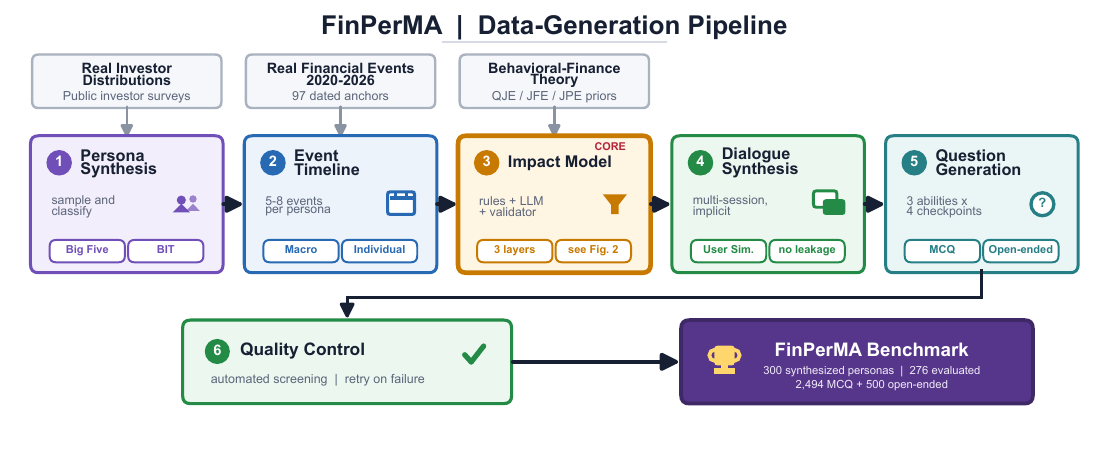}
\caption{The \finperma{} data-generation pipeline. Stage~1 samples personas from empirical distributions. Stage~2 constructs event timelines from real 2020--2026 events. Stage~3 applies the three-layer Impact Model (Figure~\ref{fig:impact}) to each $(P, E)$ pair. Stage~4 synthesizes multi-session dialogues. Stage~5 produces evaluation questions across the 3$\times$4 (dimension $\times$ checkpoint) grid.}
\label{fig:pipeline}
\end{figure*}

\paragraph{Our approach.}
We introduce \finperma{}, a controlled benchmark for event-conditioned personalized memory (Figure~\ref{fig:pipeline}). A three-layer \emph{Impact Model} operationalizes theory-informed heuristics as an auditable generation scaffold: Layer~1 deterministically constructs an \impactc{}, Layer~2 narrates a candidate response under that constraint, and Layer~3 applies implemented checks with retry. The event stream is anchored in dated 2020--2026 financial and life events, while the \emph{Post-Shock} checkpoint tests whether a memory system uses an updated rather than stale user model. The frozen v8gold corpus makes all systems face the same trajectories, so comparisons do not depend on regenerating labels for each evaluated model.

\paragraph{Contributions.} \textbf{(i)} A \textbf{dynamic personalized-memory
benchmark} that jointly tests stable-profile recall and event-conditioned
preference adaptation, separating recall from state updating. \textbf{(ii)} The
\textbf{Post-Shock checkpoint}, a direct test of whether a consequential event
has been integrated into persistent memory. \textbf{(iii)} An
\textbf{auditable generation pipeline} combining deterministic theory-informed
rules, controlled narration, and automated quality screening over 97 dated
macro, industry, and personal events. \textbf{(iv)} A \textbf{frozen empirical
evaluation} of seven LLM backbones and multiple memory architectures on
$2{,}994$ questions. Capability, checkpoint, failure-mode, and token-efficiency
analyses locate failures in recall, updating, and consolidation rather than
reducing performance to one score. We release the corpus, rule engine, model
identifiers, prompts, and seeds.

\section{Methodology}

\subsection{System Overview}
\label{sec:overview}

\finperma{} consists of five sequential stages producing a corpus
\begin{equation}
    \mathcal{C} = \{(P_i, \mathcal{T}_i, \mathcal{D}_i, \mathcal{Q}_i)\}_{i=1}^{N},
\end{equation}
where for each of $N \approx 300$ personas: $P_i$ is a persona description; $\mathcal{T}_i = (E_{i,1}, \dots, E_{i,K_i})$ is a persona-specific event timeline with $K_i \in [5, 8]$; $\mathcal{D}_i$ is the set of multi-session dialogues synthesized around the timeline; and $\mathcal{Q}_i$ is the evaluation-question set distributed across three evaluation dimensions and four temporal checkpoints. The central object of interest, computed inside Stage~3, is the \emph{event reaction} $R_{i,k}$ associated with event $E_{i,k}$:
\begin{equation}
R_{i,k} = \textsc{ImpactModel}\bigl(P_i, E_{i,k}, \mathcal{M}_{i, <k}\bigr),
\label{eq:reaction}
\end{equation}
where $\mathcal{M}_{i, <k}$ denotes persona $i$'s memory bank prior to event $k$. The Impact Model is deterministic-plus-stochastic: Layer~1 deterministically produces an \impactc{}, Layer~2 stochastically produces a candidate \eventr{} within the constraint, and Layer~3 validates and retries.

\subsection{Persona Synthesis}
\label{sec:persona}

We sample personas from a joint distribution over demographic, financial-profile, and psychometric variables. Selected demographic and financial-profile marginals, including age, education, income tier, investment experience, and product participation, are calibrated to aggregate statistics reported in publicly available retail-investor surveys conducted by the Shenzhen Stock Exchange and the Asset Management Association of China \citep{szse2020investor,amac2020investor}. Remaining attributes follow predefined sampling priors. Psychometric variables comprise (i) Big-Five personality traits, whose association with risk tolerance has been documented in individual-investor studies \citep{durand2008intrinsic}; (ii) a 10-item risk-tolerance score; and (iii) a 5-item financial literacy quiz.

Each sampled persona is then classified into one of four Behavioral Investor Types (BITs): Passive Preserver (PP), Friendly Follower (FF), Independent Individualist (II), or Active Accumulator (AA). We use a rule-based mapping motivated by \citet{pompian2012bit}. To ensure balanced coverage, rejection sampling is applied so that each BIT type occupies $\geq 20\%$ of the corpus.

\subsection{Event Timeline Construction}
\label{sec:events}

We curate 97 dated events spanning 17 macro, 64 industry, and 16 personal events. Each event records its type, date, affected assets or domains, region, summary, typical responses, and a severity score on a 1--5 ordinal scale. Rather than storing a binary per-persona exposure field, industry events are selected with a seeded relevance filter based on interest-tag overlap, knowledge domains, information-consumption style, and event salience. For each persona, we sample a timeline of $K \in [5, 8]$ events subject to category diversification and a minimum inter-event spacing of four weeks.

\subsection{The Impact Model}
\label{sec:impact}

\begin{figure*}[t]
\centering
\includegraphics[width=0.62\textwidth]{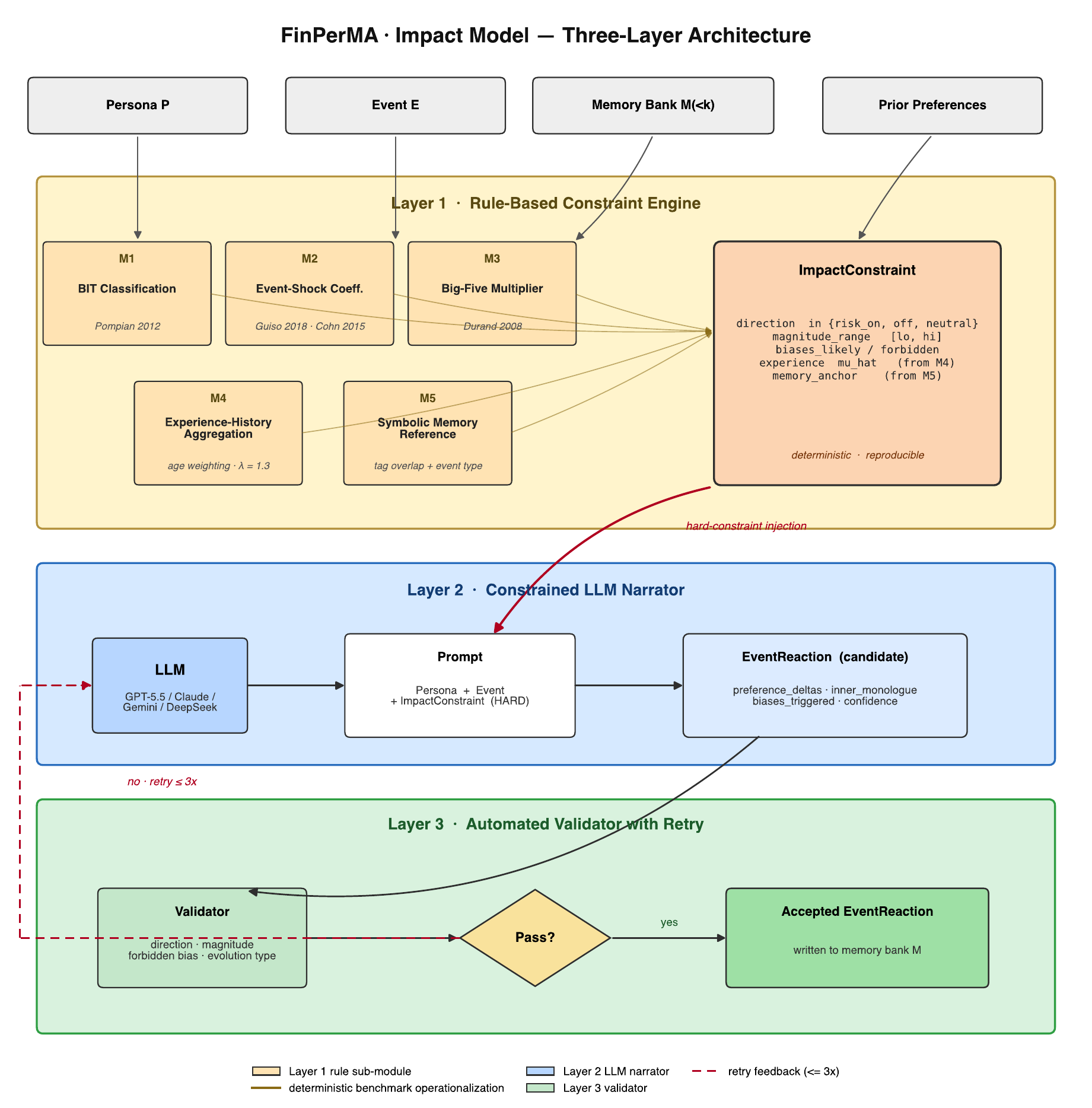}
\caption{The three-layer Impact Model. Layer~1 composes five deterministic sub-modules into an \impactc{}. Layer~2 narrates a candidate \eventr{} under structured constraints. The implemented Layer~3 hard checks cover risk direction, magnitude bounds, forbidden biases, and evolution type; failed checks trigger up to three retries. Experience and memory-reference signals are retained as soft diagnostics.}
\label{fig:impact}
\end{figure*}

The Impact Model is a transparent generation scaffold for \finperma{}. Its three-layer architecture (Figure~\ref{fig:impact}) combines (i) deterministic, theory-informed operationalizations, (ii) controlled natural-language realization, and (iii) automated validation. It is not an empirically estimated model of human behavior; its purpose is to produce reproducible, internally consistent trajectories for comparing memory systems.

\subsubsection*{Layer~1: Rule-Based \impactc{}}
Layer~1 comprises five sub-modules whose outputs are aggregated into a single \impactc{} object.

\paragraph{M1: BIT classification.} Following \citet{pompian2012bit}, we classify each persona at synthesis time. BIT type gates the direction and forbidden-bias sets used by downstream modules.

\paragraph{M2: Event-shock coefficients.} Motivated by evidence on time-varying risk response \citep{guiso2018time,cohn2015evidence}, a lookup table maps (BIT, event type) to a direction and event-type coefficient. With severity recorded on a 1--5 scale, the implementation computes
\begin{equation}
    \text{base\_mag}(E) = \mathbf{c}[E.\text{type}] \cdot \frac{E.\text{severity}}{5}.
\end{equation}
This is a deterministic benchmark operationalization rather than an empirically estimated structural coefficient.

\paragraph{M3: Big-Five trait multiplier.} Following individual-investor personality studies \citep{durand2008intrinsic}, the base magnitude is scaled by a personality-based multiplier
\begin{equation}
    m_{\text{trait}}(P) = 1 + 0.30 (P.N - 0.5) - 0.15 (P.E - 0.5),
\end{equation}
where $P.N$ and $P.E$ denote the persona's Neuroticism and Extraversion scores in $[0,1]$. Openness, Conscientiousness, and Agreeableness are omitted from the multiplier due to inconsistent effect directions in the literature.

\paragraph{M4: Experience-history aggregation.} Motivated by experience-based belief formation \citep{malmendier2011depression,malmendier2016inflation}, the released v8gold generator aggregates experienced returns with an age-dependent deterministic weight:
\begin{equation}
\begin{aligned}
 \hat{\mu}_t^{(P)} &= \sum_i w_i r_i, \\
 w_i &= \frac{a_i^{\lambda}}{\sum_j a_j^{\lambda}}, \\
 a_i &= \max(1,t-y_i+1), \qquad \lambda=1.3.
\end{aligned}
\end{equation}
The resulting belief modulates impact magnitude by $\operatorname{clip}(1-0.2\hat{\mu},0.6,1.4)$. Because this implementation increases weight with elapsed experience age, we treat it as a lifetime-experience heuristic, not as a structural recency estimate.

\paragraph{M5: Symbolic memory-reference anchor.} Motivated by memory-based reference dependence \citep{bordalo2020memory}, the implementation activates a memory when
\begin{equation}
 s(m,x)=\min\!\left(1,0.8\,J(T_m,T_x)+0.2\,\mathbb{I}[e_m=e_x]\right)\geq 0.3,
\end{equation}
where $J$ is tag-set Jaccard overlap and $e$ is event type. Activated memories receive strength $h_m=\exp(-0.1\,\text{years}_m)(1+\text{rehearsal}_m)$, and the reference value is the $s(m,x)h_m$-weighted average of stored values. This symbolic mechanism is deliberately interpretable and auditable; no embedding similarity or temperature parameter is used in v8gold.

\paragraph{Aggregation.} The five modules jointly produce the \impactc{}:
\begin{verbatim}
@dataclass
class ImpactConstraint:
    direction: Literal["risk_on",
                       "risk_off",
                       "neutral"]
    magnitude_range: Tuple[float, float]
    biases_likely: Set[BiasType]
    biases_forbidden: Set[BiasType]
    experience_mu_hat: float
    memory_anchor: List[MemoryRef]
\end{verbatim}
Given fixed $(P, E, \mathcal{M})$, this object is deterministic: the same inputs across independent runs produce the same output.

\subsubsection*{Layer~2: Constrained LLM Narrator}
Layer~2 receives the \impactc{} and produces a candidate \eventr{} via LLM completion. Following the framing of retrieval-augmented and constraint-augmented generation \citep{lewis2020rag}, we insert the constraint into the prompt as a \emph{hard} boundary rather than a soft suggestion:
\begin{quote}\small\ttfamily
[Persona] \{persona\}; [Event] \{event\}.\\
{}[Hard constraints, MUST satisfy] direction=\{dir\}; magnitude$\in$[\{lo\},\{hi\}]; allowed/forbidden biases; experience $\hat{\mu}$=\{mu\}; memory anchor.\\
{}[Output] JSON \{"preference\_deltas": ...\}
\end{quote}
The narrator produces four fields: (i) \texttt{preference\_deltas}, structured tuples of the form \texttt{(field, delta, new\_value)}; (ii) \texttt{inner\_monologue}, a natural-language rationalization that must reference at least one item in \texttt{memory\_anchor}; (iii) \texttt{biases\_triggered}, a subset of \texttt{biases\_likely}; (iv) \texttt{confidence}, a self-reported scalar in $[0, 1]$. The structured JSON schema exposes each generated field directly to the automated checks in Layer~3.

\subsubsection*{Layer~3: Automated Validator with Retry}
Layer~3 enforces the checks implemented in the released validator: risk-direction consistency; risk-delta bounds with a 20\% tolerance; exclusion of \texttt{biases\_forbidden}; and evolution-type consistency for shocks. Experience-belief and memory-reference conditions generate soft diagnostics but do not reject a candidate. Failed hard checks trigger retry feedback, up to three attempts; if all attempts fail, a template fallback is emitted with \texttt{confidence}$=0.2$ and \texttt{layer3\_fallback=True}.

\begin{algorithm}[t]
\caption{Impact Model forward pass}
\label{alg:impact}
\begin{algorithmic}[1]
\REQUIRE persona $P$, event $E$, memory bank $\mathcal{M}$
\ENSURE event reaction $R$
\STATE $\mathbf{c} \gets \textsc{ComputeConstraint}(P, E, \mathcal{M})$
\FOR{$t = 1, 2, 3$}
    \STATE $R_{\text{cand}} \gets \textsc{LLMNarrator}(P, E, \mathbf{c})$
    \IF{$\textsc{Validate}(R_{\text{cand}}, \mathbf{c})$}
        \STATE \textbf{return} $R_{\text{cand}}$
    \ELSE
        \STATE $\mathbf{c} \gets \mathbf{c} \cup \textsc{FailureFeedback}(R_{\text{cand}}, \mathbf{c})$
    \ENDIF
\ENDFOR
\STATE \textbf{return} $\textsc{Fallback}(\mathbf{c})$
\end{algorithmic}
\end{algorithm}

\subsection{Dialogue Synthesis}
\label{sec:dialogue}

For each event $E_{i,k}$ and reaction $R_{i,k}$, a Task-Oriented User Simulator receives $(P_i,E_{i,k},R_{i,k})$ as private state and generates an investor--advisor dialogue in which preferences surface implicitly through trades, questions, and concerns. Automated quality control first applies deterministic deduplication, implicitness, and financial-plausibility filters, then uses multi-LLM quality voting; only passing sessions enter the frozen corpus.

\subsection{Evaluation Task Design}
\label{sec:eval}

\begin{figure*}[t]
\centering
\includegraphics[width=0.82\textwidth]{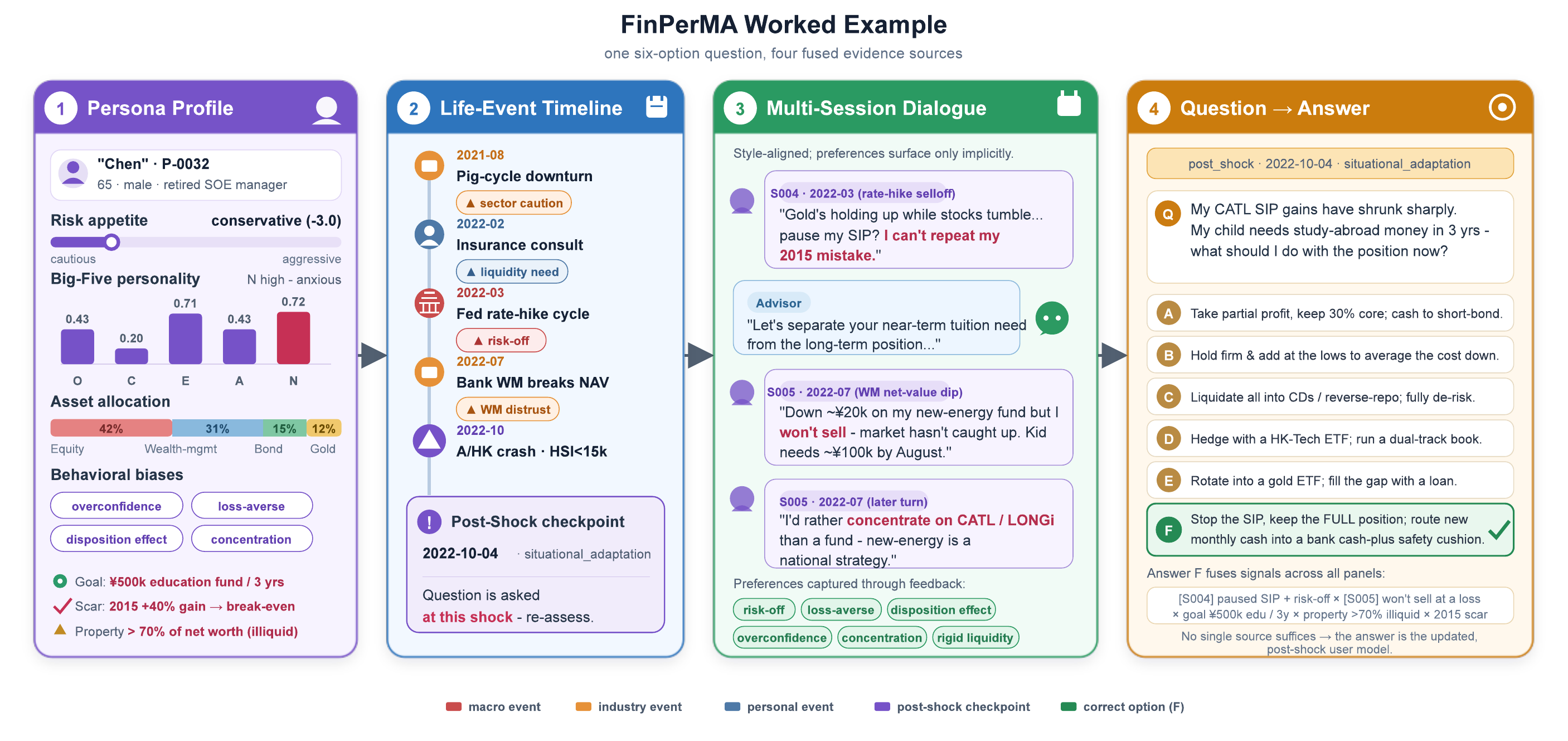}
\caption{A worked example of a single \finperma{} evaluation instance for persona P-0032. Panels~1--3 provide the persona profile, life-event timeline, and multi-session dialogue, respectively. The profile records traits and financial attributes. The timeline records macro, industry, and personal events through the Post-Shock checkpoint, while the dialogue reveals preferences implicitly. Panel~4 poses a six-option question with structurally matched distractors. Answering it requires combining all three evidence sources to infer the \emph{updated, post-shock} user model.}
\label{fig:example}
\end{figure*}

\subsubsection*{The Three-Dimensional Framework}
\finperma{} evaluates memory along three dimensions: \textbf{Memory Fidelity (MF)}, factual recall of the persona's stated attributes and past behaviors; \textbf{Preference Reasoning (PR)}, inference of implicit preferences from behavioral traces; and \textbf{Context Adaptation (CA)}, application of the (possibly updated) preferences in a novel decision scenario.

\subsubsection*{Four Temporal Checkpoints}
Orthogonal to the dimensions, evaluation runs at four checkpoints. \textbf{Zero-memory (control)} uses no persona history and measures the parametric-knowledge floor. \textbf{In-timeline} tests sustained retention as events unfold, while \textbf{Post-distraction} tests retention after unrelated sessions. \textbf{Post-Shock (novel)} is posed immediately after a high-severity event and tests whether the agent has integrated that event into a persistently updated user model.
The Post-Shock checkpoint is our principal evaluation-protocol contribution. LongMemEval tests retention and updates at the fact level \citep{wu2024longmemeval}, whereas LoCoMo tests continuity over very long conversations \citep{maharana2024locomo}. Neither isolates whether an agent integrates a consequential event into its persistent user model.

\subsubsection*{Question Generation and Distractor Design}
Figure~\ref{fig:example} shows a complete worked example of a single \finperma{} instance, in which answering one multiple-choice question requires jointly leveraging the persona profile, the life-event timeline, and the multi-session dialogue. Across the frozen corpus, personas contribute 3--19 questions (median 11) over the $3\times4$ dimension--checkpoint grid. The benchmark contains $2{,}494$ six-option MCQs (83.3\%) and 500 open-ended questions (16.7\%). MCQ answer positions are uniformly shuffled with a fixed seed at construction time. Distractors are designed for comparable structure and plausibility; empirically, the median absolute length difference from the correct option is five characters, while 52.9\% fall within five characters. We report these distributional properties rather than treating them as universal hard constraints.

\subsubsection*{Model Evaluation Protocol}
We evaluate seven frontier LLMs as backbones, each under a no-memory lower bound and a full-context upper bound. On Qwen3.7-Max we additionally compare five memory frameworks: BM25 and BGE-M3 retrieval and three structured/profile systems (Mem0 \citep{chhikara2025mem0}, MemOS \citep{li2025memos}, Memobase), all on the frozen v8gold set ($2{,}994$ questions, single seed). MCQ correctness is computed deterministically by comparing the selected option's content with the gold option after construction-time shuffling. Open-ended correctness is determined by majority vote from a three-way cross-vendor judge ensemble \citep{zheng2023judging}; judge-produced auxiliary scores are used for preference alignment, bias identification, and memory fidelity.

\begin{figure*}[t]
\centering
\includegraphics[width=0.80\textwidth]{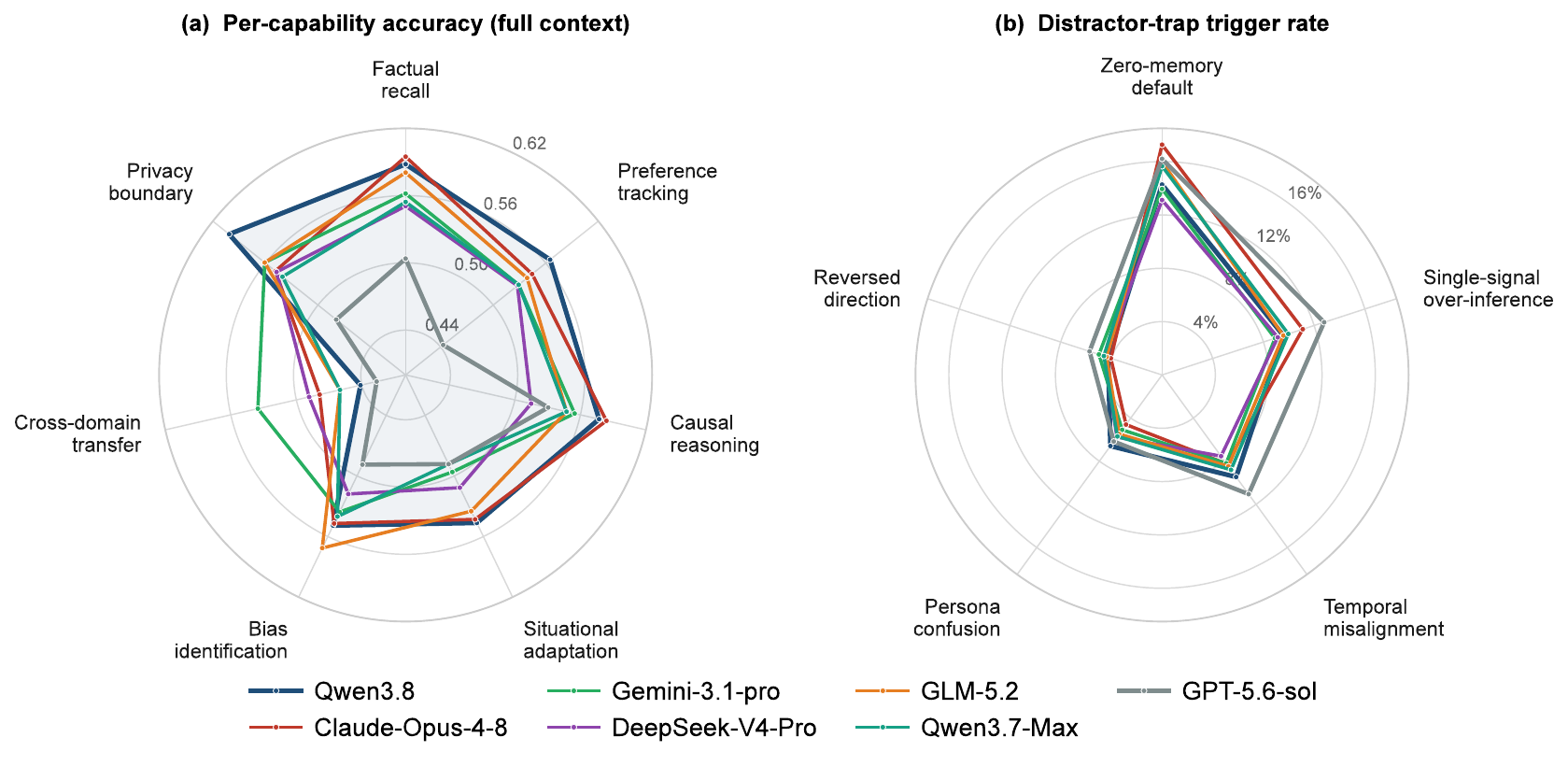}
\caption{Attribution radar under \texttt{full\_context}. \textbf{(a)} Per-capability accuracy across the seven ability axes (zero-memory control questions removed); no single backbone leads on all axes. \textbf{(b)} Distractor-trap trigger rate by cognitive-failure mode, where each error option is pre-typed by its intended failure; every model's dominant mode is \emph{zero-memory default}, and the GPT-5.6-sol is distinctively prone to \emph{single-signal over-inference} and \emph{temporal misalignment}.}
\label{fig:radar}
\end{figure*}

\paragraph{Metrics.}\looseness=-1 \emph{Acc}, \emph{MCQ}, and \emph{Open} are percent-correct overall, on MCQs, and on open-ended items. MCQ correctness uses deterministic option-content matching; open-ended correctness uses judge-majority vote. The judge ensemble also returns \emph{PAS} and \emph{BIA} in $[0,1]$ and \emph{MemFid} in $[0,5]$. \emph{\%UB} is the share of the \texttt{no\_memory}--\texttt{full\_context} accuracy gap recovered; \emph{Ctx} is mean context tokens per query, and $\eta_{1\text{k}}$ is accuracy gain over \texttt{no\_memory} per additional 1k tokens. For Fig.~\ref{fig:radar}, each MCQ has one of seven ability labels and each wrong option a pre-labeled failure mode.

\section{Evaluation}
\label{sec:exp}
We evaluate \finperma{} on 2{,}994 questions from 276 personas across seven
backbones and up to seven memory configurations, followed by attribution
analysis.

\begin{table*}[t]
\centering
\footnotesize
\setlength{\tabcolsep}{4.5pt}
\renewcommand{\arraystretch}{1.0}
\begin{tabular}{lccccccc}
\toprule
\rowcolor[gray]{0.90}
\multicolumn{8}{c}{\emph{Panel A: Base-model comparison (\texttt{no\_memory} lower bound vs.\ \texttt{full\_context} upper bound)}} \\
\midrule
Model & Acc$_\text{no-mem}$ & Acc$_\text{full}$ & MCQ & Open & PAS & BIA & MemFid \\
\midrule
Qwen-3.8         & 27.3 & \textbf{46.9} & 38.7 & \textbf{87.8} & \textbf{0.671} & \textbf{0.587} & 1.94 \\
Claude-Opus-4-8 & 31.2 & 46.6 & \textbf{39.1} & 84.3 & 0.635 & 0.570 & 1.95 \\
Gemini-3.1-pro  & 20.2 & 44.9 & 38.7 & 75.8 & 0.596 & 0.519 & 1.81 \\
DeepSeek-V4-Pro & 27.3 & 44.5 & 37.9 & 77.2 & 0.649 & 0.562 & 1.89 \\
GLM-5.2         & 21.1 & 45.6 & 39.0 & 78.2 & 0.648 & 0.555 & \textbf{1.97} \\
Qwen3.7-Max     & 18.2 & 43.6 & 38.3 & 70.0 & 0.624 & 0.539 & 1.83 \\
GPT-5.6-sol & 23.3 & 40.1 & 31.0 & 85.9 & 0.649 & 0.536 & 1.59 \\
\midrule
\rowcolor[gray]{0.90}
\multicolumn{8}{c}{\emph{Panel B: Memory-system comparison, base model fixed at Qwen3.7-Max}} \\
\midrule
Memory system & Acc & MCQ & Open & MemFid & \%\,UB & Ctx$_\text{k}$ & $\eta_{1\text{k}}$ \\
\midrule
\texttt{full\_context} (UB) & 43.6 & 38.3 & 70.0 & 1.83 & 100 & 12.8 & 0.021 \\
Bge\_rag (dense) & \textbf{40.6} & 35.6 & 65.8 & 1.62 & \textbf{88.2} & 1.40 & 0.251 \\
Naive\_rag (BM25) & 40.3 & 35.0 & \textbf{67.2} & \textbf{1.64} & 87.0 & 1.57 & 0.208 \\
Memobase & 38.3 & \textbf{35.8} & 50.8 & 1.55 & 79.1 & 1.18 & 0.301 \\
Mem0 & 36.6 & 34.0 & 49.8 & 1.53 & 72.4 & 1.02 & \textbf{0.360} \\
MemOS & 32.0 & 32.6 & 28.8 & 1.49 & 54.3 & 0.52 & -- \\
\texttt{no\_memory} (LB) & 18.2 & 17.2 & 23.6 & 0.78 & 0 & 0.51 & -- \\
\bottomrule
\end{tabular}
\caption{Evaluation results on \finperma{} ($n{=}2994$: $2494$ MCQ $+$ $500$ open-ended). See the ``Metrics '' paragraph for definitions. Panel~B swaps PAS/BIA for Ctx$_\text{k}$ (context tokens/query, thousands) and $\eta_{1\text{k}}$ (\texttt{--}: negligible budget). \textbf{Panel~A}: persona history lifts overall accuracy $1.5$--$2.4\times$, with Qwen3.8 leading. \textbf{Panel~B} (Qwen3.7-Max): retrieval recovers ${\approx}88\%$ of the gap on ${\sim}$one-tenth the tokens of \texttt{full\_context}; structured memory trails on accuracy but is most token-efficient. $^\dagger$Claude-Opus-4-8 and GPT-5.6-sol \texttt{full\_context} use $n{=}2959$ ($35$ queries dropped by the gateway content-safety filter).}
\label{tab:main}
\end{table*}

\subsection{Experimental Setup}
\label{sec:exp:setup}
We evaluate seven recent frontier LLMs representative of systems used in
Chinese-market financial-advisory settings: Qwen3.8, Qwen3.7-Max,
DeepSeek-V4-Pro, GLM-5.2, GPT-5.6-sol, Gemini-3.1-pro, and
Claude-Opus-4-8. Each uses the provider default system prompt. We probe seven
points on the memory-capability spectrum. The references are
\texttt{no\_memory}, which receives only the query, and \texttt{full\_context},
which receives all prior turns verbatim. Retrieval baselines comprise
\texttt{naive\_rag} with BM25 and \texttt{bge\_rag} with BGE-M3
\citep{lewis2020rag}. The extraction/profile systems are \texttt{mem0}
\citep{chhikara2025mem0}, \texttt{memos} \citep{li2025memos}, and
\texttt{memobase}. All backbones use the
\texttt{no\_memory} and \texttt{full\_context} references; the
seven-system comparison fixes Qwen3.7-Max. MCQ items are graded
deterministically against the reference option; open-ended items receive a
binary label and quality scores (PAS, BIA in $[0,1]$; memory fidelity in
$[0,5]$) from a cross-vendor multi-judge vote \citep{zheng2023judging}.

\subsection{Main Results}
\label{sec:exp:results}
Table~\ref{tab:main} reports the two reference points across all seven
base models (Panel~A) and the seven-way memory-system comparison on
Qwen3.7-Max (Panel~B).

\paragraph{Finding~1: A wide, model-dependent gap.}
Persona history lifts every model far above its parametric floor
(Table~\ref{tab:main}, Panel~A): overall accuracy rises from
$0.18$--$0.27$ without memory to $0.41$--$0.47$ under full context, a
$1.5$--$2.4\times$ jump. The gain is largest where the parametric
floor is weakest (Qwen3.7-Max, $2.4\times$) and smallest for the
strongest priors (Claude-Opus-4-8, $1.5\times$); Qwen3.8 leads on both
overall accuracy ($0.469$) and open-ended quality ($0.878$). The consistent gain
shows that stronger parametric knowledge does not replace persona-specific
evidence. MCQ accuracy is especially memory-dependent. Without history, it
remains near chance ($0.17$--$0.26$), and no full-context MCQ score exceeds
$\approx\!39\%$, so \finperma{} remains far from saturated.

\paragraph{Finding~2: Retrieval leads under a fixed backbone.}
\looseness=-1 On the complete Qwen3.7-Max grid (Panel~B), both retrievers recover ${\approx}88\%$ of the lower-to-upper-bound gap, while the three structured/profile memory systems recover $54$--$79\%$. The difference concentrates on open-ended questions: retrieval preserves verbatim evidence that summary/profile memory may discard. Cost changes the ranking. \texttt{bge\_rag} reaches the $88\%$ level with $1.40$k rather than $12.8$k context tokens, giving a $12\times$ marginal-efficiency edge, while \texttt{mem0} is the most token-efficient system. Under a fixed budget, lightweight retrieval can therefore be preferable to the accuracy leader. This fixed-backbone comparison uses native, unmatched token budgets, so its ranking describes operating points rather than controlled architecture quality.

\subsection{Attribution Analysis}
\label{sec:exp:attr}

\paragraph{Where and why memory fails.}
\looseness=-1 By checkpoint, the Qwen3.7-Max
\texttt{full\_context}$-$\texttt{mem0} gap widens from $8.0$ in-timeline
($53.4$ vs $45.4$) to $13.0$ at Post-Shock ($51.9$ vs $38.9$), while a
zero-memory control stays at the $\sim\!17\%$ floor. The gap widens precisely
when new evidence should revise the stored user state, making Post-Shock a
distinct test of memory updating. By capability, \texttt{memos} matches
\texttt{full\_context} on \emph{factual recall} ($27.7$ vs $27.1$) yet falls
sharply on inference targets (\emph{preference tracking} $31.9$ vs $52.9$;
\emph{bias identification} $32.6$ vs $54.0$). Consolidation therefore preserves
surface facts more reliably than the preference signals needed for
personalization. This low value is partly a control-set artifact ($76\%$ of such
MCQs are zero-memory by design); once removed, \emph{situational adaptation} and
\emph{cross-domain transfer} are the hardest capabilities
(Fig.~\ref{fig:radar}a).

\paragraph{Who fails, and how.}
\looseness=-1 No backbone leads on all seven capabilities
(Fig.~\ref{fig:radar}a): the lead rotates across Claude (factual recall, causal
reasoning), Qwen3.8 (preference tracking, situational adaptation), and Gemini
(cross-domain transfer). Overall model strength therefore does not translate
into uniformly stronger personalized memory. Pre-typed distractors expose
distinct \emph{fingerprints} (Fig.~\ref{fig:radar}b). The top error is
\emph{zero-memory default}, followed by \emph{single-signal over-inference} and
\emph{temporal misalignment}. Together, these modes account for
$\approx\!66\%$ of errors and isolate recall, integration, and recency as
separate bottlenecks. All models score $4.5$--$11.2$ points lower on
\emph{anti-typical} than typical personas, indicating a fallback to
type-level stereotypes when individual evidence conflicts with a familiar
profile \citep{jiang2025personamem}.

\subsection{Discussion}
\paragraph{What the results show.} \finperma{} separates access to user history
from the ability to maintain a changing user model. Full context improves every
backbone, confirming that persona-specific evidence matters, yet the benchmark
remains far from saturated. Under a fixed backbone, retrieval approaches the
full-context upper bound with far fewer tokens, whereas profile-based
consolidation often retains facts but loses preference cues. The challenge is
therefore not simply storing more text, but preserving evidence that can revise
the current user state.

\paragraph{Implications for memory design.} The results favor a hybrid design
that separates stable attributes from mutable preferences, timestamps changes,
and retrieves the dialogue that supports them. The Post-Shock checkpoint tests
whether this update occurs after a consequential event. The anti-typical gap
adds a second requirement: user-specific evidence must override a familiar
stereotype when the two conflict. Together, these tests distinguish persistent
personalization from factual recall and profile-based guessing.

\paragraph{Limitations and future work.} \finperma{} uses synthetic personas
and rule-guided preference changes, so benchmark accuracy does not establish
performance with real investors. Its timelines contain five to eight events and
focus on personal finance, while the memory-system comparison uses one seed per
configuration and unmatched native token budgets. Future work should add longer
histories, repeated runs, and other personalization domains. Where privacy
permits, human review and longitudinal user interactions could test whether the
generated trajectories and conclusions transfer beyond the controlled setting.

\medskip
\noindent{\small\textbf{Use of AI assistance.} LLM-based tools assisted
language editing and figure preparation. The authors reviewed all AI-assisted
material and take full responsibility for the paper.}

\clearpage

\bibliography{aaai2027}

\end{document}